\documentclass[runningheads,a4paper]{llncs}
\usepackage{multirow}
\usepackage[utf8]{inputenc} 
\usepackage[T1]{fontenc}    
\usepackage{hyperref}       
\usepackage{url}            
\usepackage{booktabs}       
\usepackage{amsfonts}       
\usepackage{nicefrac}       
\usepackage{microtype}      
\usepackage{amsmath,amssymb,amsxtra}
\usepackage{euscript}
\usepackage{dsfont}
\usepackage{graphicx}
\usepackage{algorithm}
\usepackage[noend]{algpseudocode}

\usepackage{capt-of}
\usepackage{booktabs}
\usepackage{varwidth}

\newcommand{\dd}{\mathrm{d}}
\newcommand{\RR}{\mathbb{R}}

\newcommand{\KL}{\mathrm{KL}}
\DeclareMathOperator{\diag}{diag}
\newcommand{\uptoconst}{\mathrel{\overset{\makebox[0pt]{\mbox{\normalfont\small\sffamily c}}}{=}}}

\newcommand{\der}[2]{\frac{\mathrm{d}#1}{\mathrm{d}#2}}
\newcommand{\deralt}[2]{\frac{\mathrm{d}}{\mathrm{d}#2}{#1}}

\DeclareMathOperator{\Tr}{Tr}

\begin{document}
\frontmatter          
\pagestyle{headings}  
%

\mainmatter              
\title{Bayesian Nonlinear Support Vector Machines for Big Data}
\titlerunning{Bayesian Nonlinear SVMs for Big Data}  
%
\author{
  Florian Wenzel\inst{1} \and Th\'eo Galy-Fajou\inst{1} \and Matth\"aus Deutsch\inst{2} \and Marius Kloft\inst{1}
}
\authorrunning{Florian Wenzel et al.} 
%
%
\institute{Humboldt University of Berlin, Germany \and G+J Digital Products Hamburg, Germany\\ \email{\{wenzelfl,galy,kloft\}@hu-berlin.de}, \email{mdeutsch@outlook.com}}

\maketitle              

				
\begin{abstract}
We propose a fast inference method for Bayesian nonlinear support vector machines that leverages stochastic variational inference and inducing points.
Our experiments show that the proposed method is faster than competing Bayesian approaches and scales easily to millions of data points. It provides additional features over frequentist competitors such as accurate predictive uncertainty estimates and automatic hyperparameter search.
\end{abstract}

\section{Introduction}

Statistical machine learning branches into two classic strands of research: Bayesian and frequentist.
In the classic supervised learning setting, both paradigms aim to find, based on training data, a function $f_\beta$ that predicts well on yet unseen test data.
The difference in the Bayesian and frequentist approach lies in the treatment of the parameter vector $\beta$ of this function.
In the \emph{frequentist} setting, we select the parameter $\beta$ that minimizes a certain loss given the training data,
from a restricted set $\mathcal B$ of limited complexity. 
In the \emph{Bayesian} school of thinking, we express our prior belief about the parameter, in the form of a probability distribution over
the parameter vector. 
When we observe data, we adapt our belief, resulting in a posterior distribution over $\beta$

Advantages of the Bayesian approach include automatic treatment of hyperparameters and direct 
quantification of the uncertainty\footnote{Note that frequentist approaches can also lead to other forms of uncertainty estimates, e.g. in form of confidence intervals. But since the classic SVM does not exhibit a probabilistic formulation these uncertainty estimates cannot be directly computed.} of the prediction in the form of class membership probabilities
which can be of tremendous importance in practice. As examples consider the following.
(1) We have collected blood samples of cancer patients and controls. The aim is to screen individuals that have 
increased likelihood of developing cancer.
The knowledge of the uncertainty in those predictions is invaluable to clinicians.
(2) In the domain of physics it is important to have a sense about the certainty level of predictions since it is mandatory to assert the statistical confidence in any physical variable measurement.
(3) In the general context of decision making, it is crucial that the uncertainty of the estimated outcome of an action can be reliably determined. 
 
Recently, it was shown that the support vector machine (SVM) \cite{Cortes95}---which is a classic supervised classification algorithm---
admits a Bayesian interpretation through the technique of data augmentation \cite{Polson11,Henao:2014:BNS:2968826.2969022}.
This so-called \emph{Bayesian nonlinear SVM} combines the best of both worlds: it inherits the 
geometric interpretation, its robustness against outliers, state-of-the-art accuracy \cite{Fernandez-Delgado:2014:WNH:2627435.2697065},
and theoretical error guarantees \cite{mohri2012foundations} from the frequentist formulation of the SVM,
but like Bayesian methods it also allows for flexible feature modeling, automatic hyperparameter tuning, and predictive uncertainty quantification.

However, existing inference methods for the Bayesian support vector machine (such as the expectation conditional maximization method introduced in \cite{Henao:2014:BNS:2968826.2969022}) 
scale rather poorly with the number of samples and are limited in application to datasets with thousands of data points \cite{Henao:2014:BNS:2968826.2969022}.
Based on stochastic variational inference \cite{JMLR:v14:hoffman13a} and inducing points \cite{hensman2013gaussian},
we develop in this paper a \emph{fast} and \emph{scalable} inference method for the nonlinear Bayesian SVM.

Our experiments show superior performance of our method over competing methods for uncertainty quantification of SVMs such as Platt's method \cite{Pla99}.
Furthermore, we show that our approach is faster (by one to three orders of magnitude) than the following competitors:
expectation conditional maximization (ECM) for nonlinear Bayesian SVM by \cite{Henao:2014:BNS:2968826.2969022}, 
Gaussian process classification \cite{Rasmussen:2005:GPM:1162254}, 
and the recently proposed scalable variational Gaussian process classification method \cite{Hensman2015}.
We apply our method to the domain of particle physics, namely on the SUSY dataset \cite{Baldi2014} (a standard benchmark in particle physics containing 5 million data points) where our method takes only 10 minutes to train on a single CPU machine.

Our experiments demonstrate that Bayesian inference techniques are mature enough to compete with corresponding frequentist approaches (such as nonlinear SVMs) in terms of scalability to big data, 
yet they offer additional benefits such as uncertainty estimation and automated hyperparameter search.

Our paper is structured as follows. In section~\ref{sec:related_work} we discuss related work and review the Bayesian nonlinear SVM model in section~\ref{sec:bsvm_model}.
In section~\ref{sec:inference} we propose our novel scalable inference algorithm, show how to optimize hyperparameters and obtain an approximate predictive distribution. 
We discuss also the special case of the linear SVM, for which we propose a specially tailored fast inference algorithm. 
Section~\ref{sec:experiments} concludes with experimental results.

\section{Related Work} \label{sec:related_work}

There has recently been significant interest in utilizing max-margin based discriminative Bayesian models for various applications. 
For example, \cite{Zhu14} employs a max-margin based Bayesian classification to discover latent semantic structures for topic models,
\cite{Xu13} uses a max-margin approach for efficient Bayesian matrix factorization, and \cite{Zhang14} develops a new max-margin approach to Hidden Markov models.

All these approaches apply the Bayesian reformulation of the classic SVM introduced by \cite{Polson11}. 
This model is extended by \cite{Henao:2014:BNS:2968826.2969022} to the nonlinear case.
The authors show improved accuracy compared to standard methods such as (non-Bayesian) SVMs and Gaussian process (GP) classification. 

However, the inference methods proposed in \cite{Polson11} and \cite{Henao:2014:BNS:2968826.2969022} have the drawback that they partially rely on point estimates 
of the latent variables and do not scale well to large datasets. 
In \cite{Luts:2014:MFV:2566994.2567027} the authors apply mean field variational inference to the linear case of the model,
but their proposed technique does not lead to substantial performance improvements and neglects the nonlinear model.

Uncertainty estimation for SVMs is usually done via Platt's technique \cite{Pla99}, which consists of applying a logistic regression on the function scores produced by the SVM. 
In contrast, our technique directly yields a sound predictive distribution instead of using a heuristically motivated transformation. 
We make use of the idea of inducing point GPs to develop a scalable inference method for the Bayesian nonlinear SVM. Sparse GPs using pseudo-inputs were first introduced in \cite{NIPS2005_sparseGP}. Building on this idea Hensman et al. developed a stochastic variational inference scheme for GP regression and GP classification \cite{hensman2013gaussian,Hensman2015}. We further extend this ideas to the setting of Bayesian nonlinear SVM.

\section{The Bayesian SVM Model} \label{sec:bsvm_model}

Let ${\cal D} = \left\{x_i, y_i\right\}_{i=1}^n$ be $n$ observations where $x_i \in {\RR}^d$ is a feature vector with corresponding labels $y_i \in \{-1,1\}$. 
The SVM aims to find an optimal score function $f$ by solving the following regularized risk minimization objective:
\begin{align}
	\arg \min_{f} \; \gamma R\left(f\right) + \sum_{i=1}^n \max\left(0,1-y_i f(x_i)\right)  \label{eq:SVM},
\end{align}
where $R$ is a regularizer function controlling the complexity of the decision function $f$, 
and $\gamma$ is a hyperparameter to adjust the trade-off between training error and the complexity of $f$.
The loss $\max\left(0,1-yf(x)\right)$ is called hinge loss. 
The classifier is then defined as $\text{sign}(f(x))$.

For the case of a linear decision function, i.e. $f(x)=x^T\beta$, the SVM optimization problem \eqref{eq:SVM} is equivalent to estimating the mode of a pseudo-posterior

\begin{align*}
	p(\beta | {\cal D}) \propto \prod_{i=1}^n L(y_i | x_i, \beta) p(\beta).
\end{align*}
Here $p(\beta )$ denotes a prior such that $\log p(\beta) \propto -2 \gamma R(\beta)$. 
In the following we use the prior $\beta \sim {\cal N}(0,\Sigma)$, where  $\Sigma \in {\RR} ^{d\times d}$ is a positive definite matrix.
From a frequentist SVM view, this choice generalizes the usual $L^2$-regularization
to non-isotropic regularizers.
Note that our proposed framework can be easily extended to other regularization techniques 
by adjusting the prior on $\beta$
(e.g. block $\ell_{(2,p)}$-norm regularization which is known as multiple kernel learning \cite{kloft2011lp}). 
In order to obtain a Bayesian interpretation of the SVM, we need to define a pseudolikelihood $L$ such that the following holds,
\begin{align}\label{eq:pseudolik}
	L\left(y | x, f(\cdot) \right) \propto \exp \left( -2 \max (1-y_if(x_i),0) \right).
\end{align}
By introducing latent variables $\lambda:= (\lambda_1,\dots,\lambda_n)^\top$ (data augmentation) and making use of integral identities stemming from function theory,
\cite{Polson11} show that the specification of $L$ in terms of the following marginal distribution satisfies \eqref{eq:pseudolik}:
\begin{align}
	L(y_i\vert x_i,\beta ) = \int _0^\infty \frac{1 }{\sqrt{2\pi \lambda _i }}\exp \left( -\frac{1 }{2}\frac{\left(1+\lambda _i -y_ix_i^T\beta\right)^2 }{\lambda _i}\right) \mathrm{d}\lambda _i.
    \label{eq:likelihood}
\end{align}
Writing $X\in \RR^{d\times n}$ for the matrix of data points and $Y = \diag(y)$, the full conditional distributions of this model are
\begin{align}
\begin{split}
	\beta | \lambda, \Sigma, {\cal D} &\sim {\cal N}\left( B(\lambda^{-1}+1),\, B\right) ,\\
	\lambda_i | \beta, {\cal D}_i &\sim \mathcal{GIG}\left(1/2 , 1, (1-y_i x_i^\top \beta)^2\right),
\end{split}
\label{eq:full_conditionals_lin}
\end{align}
with $Z=YX$, $B^{-1}=Z\Lambda^{-1}Z^\top + \Sigma^{-1}$, $\Lambda = \diag (\lambda )$ and where $\mathcal{GIG}$ denotes a generalized inverse Gaussian distribution. The $n$ latent variables $\lambda _i$ of the model scale the variance of the full posteriors locally. 
The model thus constitutes a special case of a normal variance-mean mixture, where we implicitly impose the improper prior $p(\lambda)=\mathds{1}_{[0,\infty)}(\lambda)$ on $\lambda$.
This could be generalized by using a generalized inverse Gaussian prior on $\lambda _i$,
leading to a conjugate model for $\lambda _i$.
Henao et al. show that in the case of an exponential prior on $\lambda _i$, this leads to a skewed Laplace full conditional for $\lambda _i$.
Note that this, however, destroys the equivalency to the frequentist linear SVM.

By using the ideas of Gaussian processes \cite{Rasmussen:2005:GPM:1162254}, Henao et al. develop a nonlinear (kernelized)  version of this model \cite{Henao:2014:BNS:2968826.2969022}. 
They assume a continuous decision function $f(x)$ to be drawn from a zero-mean Gaussian
process $\mathrm{GP}(0, k)$, where $k$ is a kernel function.
The random Gaussian vector $f = (f_1, . . . , f_n)^\top$ corresponds to $f(x)$ evaluated at the data points.
They substitute the linear function $x_i^\top \beta$ by $f_i$ in \eqref{eq:likelihood} and obtain the conditional posteriors
\begin{align}
\begin{split}
	f | \lambda, {\cal D} &\sim {\cal N}\left( CY(\lambda^{-1} +1),\, C\right) ,\\
	\lambda_i | f_i, {\cal D}_i &\sim \mathcal{GIG}\left(1/2 , 1, (1-y_if_i)^2\right),
\end{split}
    \label{eq:kernel_full_conditionals}
\end{align}
with $C^{-1} = \Lambda^{-1} + K^{-1}$. For a test point $x_*$ the conditional predictive distribution for $f_* = f(x_*)$ under this model is 
\begin{align*}
	f_* | \lambda, x_*, {\cal D} \sim {\cal N}\left(k_*^\top (K + \Lambda)^{-1}Y(1+\lambda),\, k_{**} - k_*^\top(K + \Lambda)^{-1} k_* \right),
\end{align*}
where $K:= k(X,X)$, $k_{X*}:=k(X,x_*)$, $k_{**}:= k(x_*,x_*)$. The conditional class membership probability is
\begin{align*}
	p (y_* = 1 | \lambda, x_*, {\cal D}) = \Phi \left( \frac{k_*^T(K+\Lambda)^{-1}Y(1+\lambda)}{1+ k_{**} - k_*^\top(K + \Lambda)^{-1} k_*}\right),
\end{align*}
where $\Phi(.)$ is the probit link function.

Note that the conditional posteriors as well as the class membership probability still depend on the local latent variables $\lambda _i$. 
We are interested in the marginal predictive distributions, but unfortunately the latent variables cannot be integrated out analytically. Both \cite{Polson11} and \cite{Henao:2014:BNS:2968826.2969022} propose MCMC-algorithms and stepwise inference schemes similar to EM-algorithms to overcome this problem. 
These methods do not scale well to big data problems and the probability estimation still relies on point estimates of the $n$-dimensional $\lambda$.
We overcome these problems proposing a scalable inference method and obtaining approximate marginal predictive distributions (that are not conditioned on $\lambda$).

\section{Scalable Inference and Automated Hyperparameter Tuning} \label{sec:inference}
In the following we develop a fast and reliable inference method for the Bayesian nonlinear SVM. 
Our method builds on the idea of using inducing points for Gaussian Processes in a stochastic variational inference setting \cite{hensman2013gaussian} that scales easily to millions of data points. 
We proceed by first discussing a standard batch variational scheme in section~\ref{sec:batchvi_non-linear} and then in section~\ref{sec:svi_non-linear} we develop our fast and scalable inference method. We show how to  automatically tune hyperparameters in section~\ref{sec:autotuning} and obtain uncertainty estimates for predictions in section~\ref{sec:pred_distribution}. Finally, we discuss the special case of the Bayesian linear SVM in section~\ref{sec:linear_bsvm}.

\subsection{Batch Variational Inference} \label{sec:batchvi_non-linear}
The idea of variational inference is to approximate the typically intractable posterior of a probabilistic model by a variational (typically factorized) distribution. We find the optimal approximating distribution by maximizing a lower bound on the evidence (the so-called ELBO) with respect to the parameters of the variational distribution, which is equivalent to minimizing the Kullback-Leibler divergence between the variational distribution and the posterior \cite{Jordan:1999:IVM:339248.339252,Wainwright:2008:GME:1498840.1498841}.

In this section we first develop a batch variational inference scheme \cite{Jordan:1999:IVM:339248.339252,Wainwright:2008:GME:1498840.1498841}, which uses the full dataset in every iteration.
We follow the structured mean field approach and choose the variational distributions within the same families as the full conditional distributions
$q(f,\lambda) = q(f) \prod_{i=1}^n q(\lambda_i)$,
with
$q(f) \equiv  {\cal N}(\mu, \zeta) \text{ and }	q(\lambda_i) \equiv \mathcal{GIG}(1/2,1,\alpha_i)$.
The coordinate ascent updates can be computed by the expected natural parameters of the corresponding full conditionals \eqref{eq:kernel_full_conditionals} leading to
\begin{align*}
	\alpha_i &= \mathbb{E}_{q(f)} [(1- y_i f_i)^2] = (1- y_i^\top \mu)^2 + y_i^\top \zeta y_i,\\
	\zeta &= \mathbb{E}_{q(\lambda)}[\left( \Lambda^{-1} + K^{-1}\right)^{-1}] = \left(A^{-\frac{1}{2}} + K^{-1}\right)^{-1},\\
	\mu &= \zeta \mathbb{E}_{q(\lambda)}[Y(\lambda^{-1} + 1)] = \zeta Y(\alpha^{-\frac{1}{2}} + 1).
\end{align*}
This concludes the batch variational inference scheme.

The downside of this approach is that it does not scale to big datasets. The covariance matrix of the variational distribution $q(f)$ has dimension $n \times n$ and has to be updated and inverted at every inference step. This operation exhibits the computational complexity ${\cal O}(n^3)$, where $n$ is the number of data points. Furthermore, in this setup we cannot apply stochastic gradient descent. We show how to overcome both problems in the next section paving the way to perform inference on big datasets.

\subsection{Stochastic Variational Inference Using Inducing Points} \label{sec:svi_non-linear}

We aim to develop a stochastic variational inference (SVI) scheme using only minibatches of the data in each iteration.
The Bayesian nonlinear SVM model does not exhibit a set of global variables. Both the number of latent variables $\lambda$ and the observations of the latent GP $f$ grow with number of data points (c.f. eq.\ref{eq:kernel_full_conditionals}), i.e. they are local variables. This hinders us from directly developing a SVI scheme. We make use of the concept of inducing points \cite{hensman2013gaussian} imposing a sparse GP acting as global variable. This allows us to apply SVI and reduces the complexity to ${\cal O}(m^3)$, where $m$ is the number of inducing points, which is independent of the number of data points.

We augment our original model \eqref{eq:kernel_full_conditionals} with $m<n$ inducing points. Let $u\in \RR^m$ be pseudo observations at inducing locations $\{\hat x_1,...,\hat x_m \}$.
We employ a prior on the inducing points, $p(u) = \mathcal N(0,K_{mm})$ and connect $f$ and $u$ setting
\begin{equation}
    p(f|u) = \mathcal N(K_{nm}K_{mm}^{-1}u, \widetilde K)
    \label{eq:lowrank_approx}
\end{equation}
where $K_{mm}$ is the kernel matrix resulting from evaluating the kernel function between all inducing points locations,
$K_{nm}$ is the cross-covariance between the data points and the inducing points and $\widetilde K$ is given by $\widetilde K = K_{nn} - K_{nm}K_{mm}^{-1}K_{mn}$.
The augmented model exhibits the joint distribution
\begin{align*}
	p(y,u,f,\lambda) = p(y, \lambda|f) p(f | u) p(u).
\end{align*}
Note that we can recover the original joint distribution by marginalizing over $u$. 
We now aim to apply the methodology of variational inference to the marginal joint distribution $p(y,u,\lambda)=\int p(y,u,f,\lambda)\dd f$.
We impose a variational distribution $q(u) = \mathcal N(u|\mu, \zeta)$ on the inducing points $u$. We follow \cite{hensman2013gaussian} and apply Jensen's inequality to obtain a lower bond on the following intractable conditional probability,
\begin{align*}
	\log p(y,\lambda|u) &= \log \mathbb{E}_{p(f|u)}\left[ p(y,\lambda |f ) \right] \\
		& \geq \mathbb{E}_{p(f|u)}\left[\log p(y,\lambda|f)\right]\\
		&= \sum_{i=1}^n\mathbb{E}_{p(f_i|u)}\left[\log p(y_i,\lambda_i|f_i)\right]\\
		&= \sum_{i=1}^n\mathbb{E}_{p(f_i|u)}\left[\log\left((2\pi\lambda_i)^{-\frac{1}{2}}\exp\left(-\frac{1}{2}\frac{(1+\lambda_i-y_if_i)^2}{\lambda_i}\right)\right)\right]
\end{align*}
\begin{align*}
		&\uptoconst -\frac{1}{2}\sum_{i=1}^n\mathbb{E}_{p(f_i|u)}\left[\log\lambda_i + \frac{(1+\lambda_i-y_if_i)^2}{\lambda_i}\right]\\
		&= -\frac{1}{2}\sum_{i=1}^n\left(\log\lambda_i  + \frac{1}{\lambda_i}\mathbb{E}_{p(f_i|u)}\left[(1+\lambda_i-y_if_i)^2\right]\right)\\
		&= -\frac{1}{2}\sum_{i=1}^n\left(\log\lambda_i  + \frac{1}{\lambda_i}\left( \widetilde{K}_{ii}  + \left(1+\lambda_i - y_i K_{im}K_{mm}^{-1}u\right)^2\right)\right)\\
		&=:\mathcal{L}_1.
\end{align*}
Plugging the lower bound $\mathcal{L}_1$ into the standard evidence lower bound (ELBO) \cite{Jordan:1999:IVM:339248.339252} leads to the new variational objective
\begin{align}
	\log p(y) &\geq \mathbb{E}_q\left[\log p(y,\lambda,u)\right] - \mathbb{E}_q\left[\log q(\lambda,u) \right]\nonumber\\
			&= \mathbb{E}_q\left[\log p(y,\lambda|u) \right] + \mathbb{E}_q\left[\log p(u)\right] - \mathbb{E}_q\left[\log q(\lambda,u) \right]\nonumber\\
			&\geq \mathbb{E}_q\left[\mathcal{L}_1\right] +  \mathbb{E}_q\left[\log p(u)\right] - \mathbb{E}_q\left[\log q(\lambda,u)\right]\label{eq:lowerbound_elbo}\\
			&= -\frac{1}{2}\sum_{i=1}^n\mathbb{E}_q\left[\log\lambda_i  + \frac{1}{\lambda_i}\left( \widetilde{K}_{ii}  + \left(1+\lambda_i - y_i K_{im}K_{mm}^{-1}u\right)^2\right)\right]\nonumber\\
			&\quad- \KL\left(q(u)||p(u)\right)  - \mathbb{E}_{q(\lambda)}\left[\log q(\lambda)\right]\nonumber\\
			&=: \mathcal{L}.\nonumber
\end{align}
The expectations can be computed analytically (details are given in the appendix) and we obtain ${\cal L}$ in closed form,
\begin{align}
	\mathcal{L}
		&\uptoconst \frac{1}{2} \log |\zeta| - \frac{1}{2}\text{tr}(K_{mm}^{-1}\zeta)-\frac{1}{2}\mu^\top K_{mm}^{-1}\mu + y^\top\kappa\mu\nonumber \\
		&\quad + \sum_{i=1}^n\left\{\log(\mathrm{B}_{\frac{1}{4}}(\sqrt{\alpha_i})) + \frac{1}{2}\log(\alpha_i)\right\}\label{eq:ELBO_nonlin}\\
        &\quad -\sum_{i=1}^n\frac{1}{2}\alpha_i^{-\frac{1}{2}}\left( 1 - \alpha_i - 2y_i\kappa_{i.}\mu + \left(\kappa(\mu\mu^\top + \zeta) \kappa^\top + \widetilde{K}\right)_{ii}  \right),\nonumber 
\end{align}
where $\kappa=K_{nm}K_{mm}^{-1}$ and $\mathrm{B}_{\frac{1}{2}}(.)$ is the modified Bessel function with parameter $\frac{1}{2}$ \cite{jorgensen2012statistical}.
This objective is amenable to stochastic optimization where we subsample from the sum to obtain a noisy gradient estimate.
We develop a stochastic variational inference scheme by following noisy natural gradients of the variational objective $\mathcal{L}$.
Using the natural gradient over the standard euclidean gradient is often favorable since
natural gradients are invariant to reparameterization of the variational family \cite{infogeom,naturalgrad} and provide effective second-order optimization updates \cite{amari98natural,JMLR:v14:hoffman13a}.
The natural gradients of $\mathcal{L}$ w.r.t. the Gaussian natural parameters $\eta_1 = \zeta^{-1}\mu$, $\eta_2 = -\frac{1}{2}\zeta^{-1}$ are
\begin{align}
	\widetilde{\nabla}_{\eta_1} \cal{L} &= \kappa^\top Y(\alpha^{-\frac{1}{2}}+1) - \eta_1 \label{eq:derivative_nonlinelbo_eta1} \\
	\widetilde{\nabla}_{\eta_2} \cal{L} &=  -\frac{1}{2}(K_{mm}^{-1} + \kappa^\top A^{-\frac{1}{2}} \kappa) - \eta_2,  \label{eq:derivative_nonlinelbo_eta2}
\end{align}
with $A = \diag(\alpha)$. Details can be found in the appendix.
The natural gradient updates always lead to a positive definite covariance matrix\footnote{This follows directly since $K_{mm}$ and $A^{-\frac{1}{2}}$ are positive definite.}
and in our implementation $\zeta$ has not to be parametrized in any way to ensure positive-definiteness.
The derivative of $\cal{L}$ w.r.t. $\alpha_i$ is
\begin{align}
\nabla_\alpha \cal{L} &= \frac{(1-y_i\kappa_i\mu)^2 + y_i(\kappa_i\zeta\kappa_i^\top + \widetilde{K}_{ii})y_i}{4\sqrt{\alpha_i}^3} - \frac{1}{4\sqrt{\alpha_i}}. \label{eq:derivative_nonlinelbo_alpha}
\end{align}
Setting it to zero gives the coordinate ascent update for $\alpha_i$,
\begin{align*}
	\alpha_i = (1-y_i\kappa_i\mu)^2 + y_i(\kappa_i\zeta\kappa_i^\top + \widetilde{K}_{ii})y_i.
\end{align*}
Details can be found in the appendix. The inducing point locations can be either treated as hyperparameters and optimized while training \cite{Titsias09variationallearning} or can be fixed before optimizing the variational objective.
We follow the first approach which is often preferred in a stochastic variational inference setup \cite{hensman2013gaussian,Hensman2015}.
The inducing point locations can be either randomly chosen as subset of the training set or via a density estimator. In our experiments we have observed that the $k$-means clustering algorithm (kMeans) \cite{Murphy:2012:MLP:2380985} yields the best results.
Combining our results, we obtain a fast stochastic variational inference algorithm for the Bayesian nonlinear SVM which is outlined in alg.~\ref{algo:SVI_non-linear}. We apply the adaptive learning rate method described in \cite{adaptiveSVI}.

\begin{algorithm} 
	\caption{Inducing Point SVI}
	\begin{algorithmic}[1] 
	\State set the learning rate schedule $\rho_t$ appropriately
	\State initialize $\eta_1$, $\eta_2$
    \State select $m$ inducing points locations (e.g. via kMeans)
    \State compute kernel matrices $K^{-1}_{mm}$ and $\widetilde{K} = K_{nn} - K_{nm}K_{mm}^{-1}K_{mn}$
	\While {not converged}
	\State get ${\cal S}$ = minibatch index set of size s
	\State update $\alpha _i=  (1-y_i\kappa_i\mu)^2 + y_i(\kappa_i\zeta\kappa_i^\top + \widetilde{K}_{ii})y_i$
	\State compute $A_{\cal S}  = \diag (\alpha_i,\;\; i \in {\cal S})$
	\State compute $\hat \eta_1 =  \kappa^\top Y(\alpha^{-\frac{1}{2}}+1)$
	\State compute $\hat \eta_2 = -\frac{1}{2}(K_{mm}^{-1} + \kappa^\top A^{-\frac{1}{2}} \kappa)$
	\State update $\eta_1 = (1-\rho_t)\eta_1 + \rho_t \hat \eta_1$
	\State update $\eta_2 = (1-\rho_t)\eta_2 + \rho_t \hat \eta_2$
	\State compute $\zeta = -\frac{1}{2}\eta^{-1}_2$
	\State compute $\mu = \zeta \eta_1$
	\EndWhile
	\State \textbf{return } $\alpha _1,\dots,\alpha _n, \mu , \zeta$
	\end{algorithmic}
	\label{algo:SVI_non-linear}
\end{algorithm}

\subsection{Auto Tuning of Hyperparameters} \label{sec:autotuning}
The probabilistic formulation of the SVM lets us directly learn the hyperparameters while training.
To this end we maximize the marginal likelihood $p(y|X,h)$, where $h$ denotes the set of hyperparameters (this approach is called empirical Bayes \cite{EB89}). We follow an approximate approach and optimize the fitted variational lower bound ${\cal L}(h)$ over $h$ by alternating between optimization steps w.r.t. the variational parameters and the hyperparameters \cite{cSGD}.
We include a gradient ascent step w.r.t. $h$ after multiple variational updates in the SVI scheme, this is commonly known as Type II maximum likelihood (ML-II) \cite{Rasmussen:2005:GPM:1162254}
\begin{align}
	\label{eq:hyperparam}
	h^{(t)} = h^{(t-1)} + \widetilde \rho_t \nabla_h {\cal L}(\alpha^{(t-1)} , \mu^{(t-1)}  , \zeta^{(t-1)} , h).
\end{align}
Since the standard SVM does not exhibit a probabilistic formulation, the hyperparameters have to be tuned via computationally very expensive methods as grid search and cross validation.
Our approach allows us to estimate the hyperparameters during training time and lets us follow gradients instead of only evaluating single hyperparameters.

In the appendix we provide the gradient of the variational objective ${\cal L}$ w.r.t. to a general kernel and show how to optimize arbitrary differentiable hyperparameters.
Our experiments exemplify our automated hyperparameter tuning approach by optimizing the hyper parameter of an RBF kernel.

\subsection{Uncertainty Predictions} \label{sec:pred_distribution}
Besides the advantage of automated hyperparameter tuning, the probabilistic formulation of the SVM leads directly to uncertainty estimates of the predictions. The standard SVM lacks this capability, and only heuristic approaches as e.g. Platt \cite{Pla99} exist. Using the approximate posterior $q(u| {\cal D}) = {\cal N}(u | \mu, \zeta)$ obtained by our stochastic variational inference method (alg.~\ref{algo:SVI_non-linear}) we compute the class membership probability for a test point $x^*$,
\begin{align*}
	p(f^*|x^*, {\cal D}) &= \int p(y^*|u, x^*)p(u|{\cal D})\dd u\\
			&\approx \int p(y^*|u, x^*)q(u|{\cal D})\dd u\\
			&= {\cal N}\left(y^* | K_{*m}K_{mm}^{-1}m, \; K_{**} - K_{*m}K_{mm}^{-1}(K_{m*} + \zeta K_{mm}^{-1}K_{m*})\right)\\
			&=: q(f^*|x^*, {\cal D}),
\end{align*}
where $K_{*m}$ denotes the kernel matrix between test and inducing points and $K_{**}$ the kernel matrix between test points. This leads to the approximate class membership distribution 
\begin{align}
	q(y^*|x^*, {\cal D}) &= \Phi\left(\frac{K_{*m}K_{mm}^{-1}m}{K_{**} - K_{*m}K_{mm}^{-1}(K_{m*} + \zeta K_{mm}^{-1}K_{m*}) + 1}  \right) \label{eq:pred_distr_nonlin}
\end{align}
where $\Phi(.)$ is the probit link function. Note that we already computed inverse $K_{mm}^{-1}$ for the training procedure leading to a computational overhead stemming only from simple matrix multiplication. Our experiments show that \eqref{eq:pred_distr_nonlin} leads to reasonable uncertainty estimates.


\subsection{Special Case of Linear Bayesian SVM} \label{sec:linear_bsvm}
We now consider the special case of using a linear kernel. If we are interested in this case we may consider the Bayesian model for the linear SVM proposed by  Polson et al. (c.f. eq.~\ref{eq:full_conditionals_lin}).
This can be favorable over using the nonlinear version since this model is formulated in primal space and, therefore, the computational complexity depends on the dimension $d$ and not on the number of data points $n$.
Furthermore, focusing directly on the linear model allows us to optimize the true ELBO,
$\mathbb{E}_q\left[\log p(y,\lambda,\beta)\right] - \mathbb{E}_q\left[\log q(\lambda,\beta) \right]$,
without the need of relying on a lower bound (as in eq.~\ref{eq:lowerbound_elbo}). This typically leads to a better approximate posterior.


We again follow the structured mean field approach and chose our variational distributions to be in the same families as the full conditionals \eqref{eq:full_conditionals_lin},
\begin{align*}
	q(\lambda _i) &\equiv \mathcal{GIG}(\frac{1}{2}, 1, \alpha_i ) \text{ and } q(\beta) \equiv \mathcal{N}(\mu, \zeta).
\end{align*}	
We use again the fact that the coordinate updates of the variational parameters can be obtained by computing the expected natural parameters of the corresponding full conditionals \eqref{eq:full_conditionals_lin} and obtain
\begin{align}
	\alpha_i &= (1-z_i^T\mu)^2 + z_i^T\zeta z_i\nonumber\\
	\zeta&= (Z A^{-\frac{1}{2}} Z^T+\Sigma ^{-1} )^{-1}\label{eq:updates_linSVM}\\
	\mu &=  \zeta Z (\alpha^{-\frac{1}{2}} + 1),\nonumber
\end{align}
where $\alpha = (\alpha _i)_{1\leq i\leq n}$, $A=\diag (\alpha)$ and $Z=YX$. Since the Bayesian Linear SVM model exhibits global and local variables we can directly employ stochastic variational inference by subsampling the data and only updating minibatches of $\alpha$.
Note that for the linear case the covariance matrices have size $d \times d$, i.e. being independent of the number of data points. Therefore, the SVI algorithm \eqref{eq:updates_linSVM} for the Bayesian Linear SVM exhibits the computational complexity ${\cal O}(d^3)$. Luts et. al develop a batch variational inference scheme for the Bayesian linear SVM but do not scale to big datasets.

The hyperparameter can be tuned analogously to \eqref{eq:hyperparam}. The class membership probabilities are
\begin{align*}
p(y_* = 1 | x^*, {\cal D}) 
		\approx \int \Phi(f_*) p(f_*| f, x^*) q(f|{\cal D}) \dd f \dd f_*
		= \Phi\left( \frac{x_*^\top \mu}{x_*^\top \zeta x_*+1} \right),
\end{align*}
where $x_*$ are the test points and $q(f|{\cal D}) = {\cal N}(f | \mu, \zeta)$ the approximate posterior obtained by the above described SVI scheme.

\section{Experiments} \label{sec:experiments}
We compare our approach against the expectation conditional maximization (ECM) method proposed by Henao et. al \cite{Henao:2014:BNS:2968826.2969022}, Gaussian process classification (GPC) \cite{Rasmussen:2005:GPM:1162254}, its recently proposed scalable stochastic variational inference version (S-GPC) \cite{Hensman2015}, and libSVM with Platt scaling \cite{CC01a,Pla99} (SVM + Platt).
For all experiments we use an RBF kernel\footnote{The RBF kernel is defined as $k(x_1,x_2,\theta)=\exp\left(-\frac{||x_1-x_2||}{\theta^2}\right)$, where $\theta$ is the length scale parameter.} with length-scale parameter $\theta$.
We perform all experiments using only one CPU core with 
2.9 GHz and 386 GB RAM.\\
Code is available at \url{github.com/theogf/BayesianSVM}.

\subsection{Prediction Performance and Uncertainty Estimation}
We experiment on seven real-world datasets and compare the prediction performance, the quality of the uncertainty estimates and run time of the methods. The results are presented in table~\ref{tab:uncertainty}.
We show that our method (S-BSVM) is up to 22 times faster than the direct competitor ECM
and up to 700 times faster than Gaussian process classification\footnote{For a comparison with the stochastic variational inference version of GPC, see section~\ref{sec:exp_runtime}.}
while outperforming the competitors in terms of prediction performance and quality of uncertainty estimates in most cases.
The non-probabilistic SVM is naturally the fastest method. Combined with the heuristic Platt scaling approach it leads to class membership probabilities but, however, still lacks the advantages of a probabilistic model (as e.g. uncertainty quantification of the learned parameters and automatic hyperparameter tuning).

To evaluate the quality of the uncertainty estimates we compute the Brier score which is considered as a good performance measure for probabilistic predictions \cite{brier1950verification} being defined as $BS = \frac{1}{n} \sum_{i=1}^N \left( y_i - q(x_i) \right)^2$, where $y_i \in \{0,1\}$ 
is the observed output and $q(x_i) \in [0,1]$ is the predicted class membership probability. 
Note that smaller Brier score indicates better performance.

The datasets are all from the R\"{a}tsch benchmark datasets \cite{Diethe15} commonly used to test the accuracy of binary nonlinear classifiers.
We perform a 10-fold cross-validation and use an RBF kernel with fixed parameters for all methods. For S-BSVM we choose the number of inducing points as $20\%$ of the training set size, except for the datasets \emph{Splice}, \emph{German} and \emph{Waveform} where we use 100 inducing points. For each dataset minibatches of 10 samples are used.

\begin{table}\centering
\begin{tabular}{l|l|l|l|l|l|l|l}
Dataset & n & dim. & &  \textbf{S-BSVM} & ECM & GPC & SVM + Platt\\
\hline
\multirow{2}{*}{Breast} & \multirow{3}{*}{263} & \multirow{3}{*}{9} & Error  & \boldmath$ .26 \pm .07 $ & $ .27 \pm .10 $ & $ .27 \pm .07 $ & $ .27 \pm .09 $\\
\multirow{2}{*}{Cancer} & & & Brier Score & \boldmath$ .18 \pm .03 $ & $ .19 \pm .05 $ & \boldmath$ .18 \pm .03 $ & $ .19 \pm .04 $\\
 & & & Time [s] &  $ 0.32 $  &  $ 1.4 $  &  $ 6.7 $  & \boldmath $ 0.04 $ \\ \hline
\multirow{3}{*}{Diabetes} & \multirow{3}{*}{768} & \multirow{3}{*}{8} & Error  & \boldmath$ .22 \pm .06 $ & $ .25 \pm .07 $ & $ .23 \pm .07 $ & $ .24 \pm .07 $\\
 & & & Brier Score & $ .16 \pm .04 $ & $ .17 \pm .04 $ & \boldmath$ .15 \pm .04 $ & $ .16 \pm .04 $\\
 & & & Time [s] &  $ 3.9 $  &  $ 33 $  &  $ 67 $  & \boldmath $ 0.11 $ \\ \hline
\multirow{3}{*}{Flare} & \multirow{3}{*}{144} & \multirow{3}{*}{9} & Error  & \boldmath$ .36 \pm .12 $ & \boldmath$ .36 \pm .12 $ & \boldmath$ .36 \pm .11 $ & \boldmath$ .36 \pm .12 $\\
 & & & Brier Score & \boldmath$ .22 \pm .05 $ & $ .25 \pm .07 $ & $ .24 \pm .03 $ & $ .24 \pm .04 $\\
 & & & Time [s] &  $ 0.08 $  &  $ 0.26 $  &  $ 1.8 $  & \boldmath $ 0.01 $ \\ \hline
\multirow{3}{*}{German} & \multirow{3}{*}{1000} & \multirow{3}{*}{20} & Error  & \boldmath$ .24 \pm .11 $ & $ .25 \pm .12 $ & $ .25 \pm .13 $ & $ .27 \pm .10 $\\
 & & & Brier Score & \boldmath$ .17 \pm .06 $ & \boldmath$ .17 \pm .05 $ & \boldmath$ .17 \pm .06 $ & $ .18 \pm .05 $\\
 & & & Time [s] &  $ 12 $  &  $ 80 $  &  $ 115 $  & \boldmath $ 0.15 $ \\ \hline
\multirow{3}{*}{Heart} & \multirow{3}{*}{270} & \multirow{3}{*}{13} & Error  & \boldmath$ .16 \pm .06 $ & $ .19 \pm .09 $ & \boldmath$ .16 \pm .06 $ & $ .17 \pm .07 $\\
 & & & Brier Score & $ .13 \pm .04 $ & $ .14 \pm .04 $ & \boldmath$ .12 \pm .03 $ & \boldmath$ .12 \pm .04 $\\
 & & & Time [s] &  $ 0.34 $  &  $ 2.2 $  &  $ 6 $  & \boldmath $ 0.04 $ \\ \hline
\multirow{3}{*}{Splice} & \multirow{3}{*}{2991} & \multirow{3}{*}{60} & Error  & $ .13 \pm .03 $ & \boldmath$ .11 \pm .03 $ & $ .32 \pm .14 $ & $ .14 \pm .01 $\\
 & & & Brier Score & $ .17 \pm .01 $ & $ .18 \pm .01 $ & $ .40 \pm .14 $ & \boldmath$ .11 \pm .01 $\\
 & & & Time [s] &  $ 18 $  &  $ 406 $  &  $ 419 $  & \boldmath $ 1.3 $ \\ \hline
\multirow{3}{*}{Waveform} & \multirow{3}{*}{5000} & \multirow{3}{*}{21} & Error  & \boldmath$ .09 \pm .02 $ & $ .10 \pm .02 $ & $ .10 \pm .02 $ & $ .10 \pm .02 $\\
 & & & Brier Score & \boldmath$ .06 \pm .01 $ & $ .15 \pm .01 $ & \boldmath$ .06 \pm .01 $ & \boldmath$ .06 \pm .01 $\\
 & & & Time [s] &  $ 12.5 $  &  $ 264 $  &  $ 8691 $  & \boldmath $ 2.3 $ \\ \hline \end{tabular}
\caption{Average prediction error and Brier score with one standard deviation.}
\label{tab:uncertainty}
\end{table}

\subsection{Big Data Experiments}
We demonstrate the scalability of our method on the SUSY dataset \cite{Baldi2014} containing 5 million points with 17 features.
This dataset size is very common in particle physics due to the simplicity of artificially generating new events as well as the quantity of data coming from particle detectors.
Since it is important to have a sense of the confidence of the predictions for such datasets the Bayesian SVM is an appropriate choice.
We use an RBF kernel\footnote{The length scale parameter tuning is not included in the training time. We found $\theta = 5.0$ by our proposed automatic tuning approach.}, 64 inducing points and minibatches of 100 points. The training of our model takes only 10 minutes without any parallelization. We use the the area under the receiver operating characteristic (ROC) curve (AUC) as performance measure since it is a standard evaluation measure on this dataset \cite{Baldi2014}.

Our method achieves an AUC of $0.84$ and a Brier score of $0.22$, whereby the state-of-the-art obtains an AUC of $0.88$ using a deep neural network (5 layers, 300 hidden units each) \cite{Baldi2014}. Note that this approach takes much longer to train and does not include uncertainty estimates. 

\subsection{Run Time} \label{sec:exp_runtime}

\begin{figure}
\centering
\includegraphics[scale=0.3]{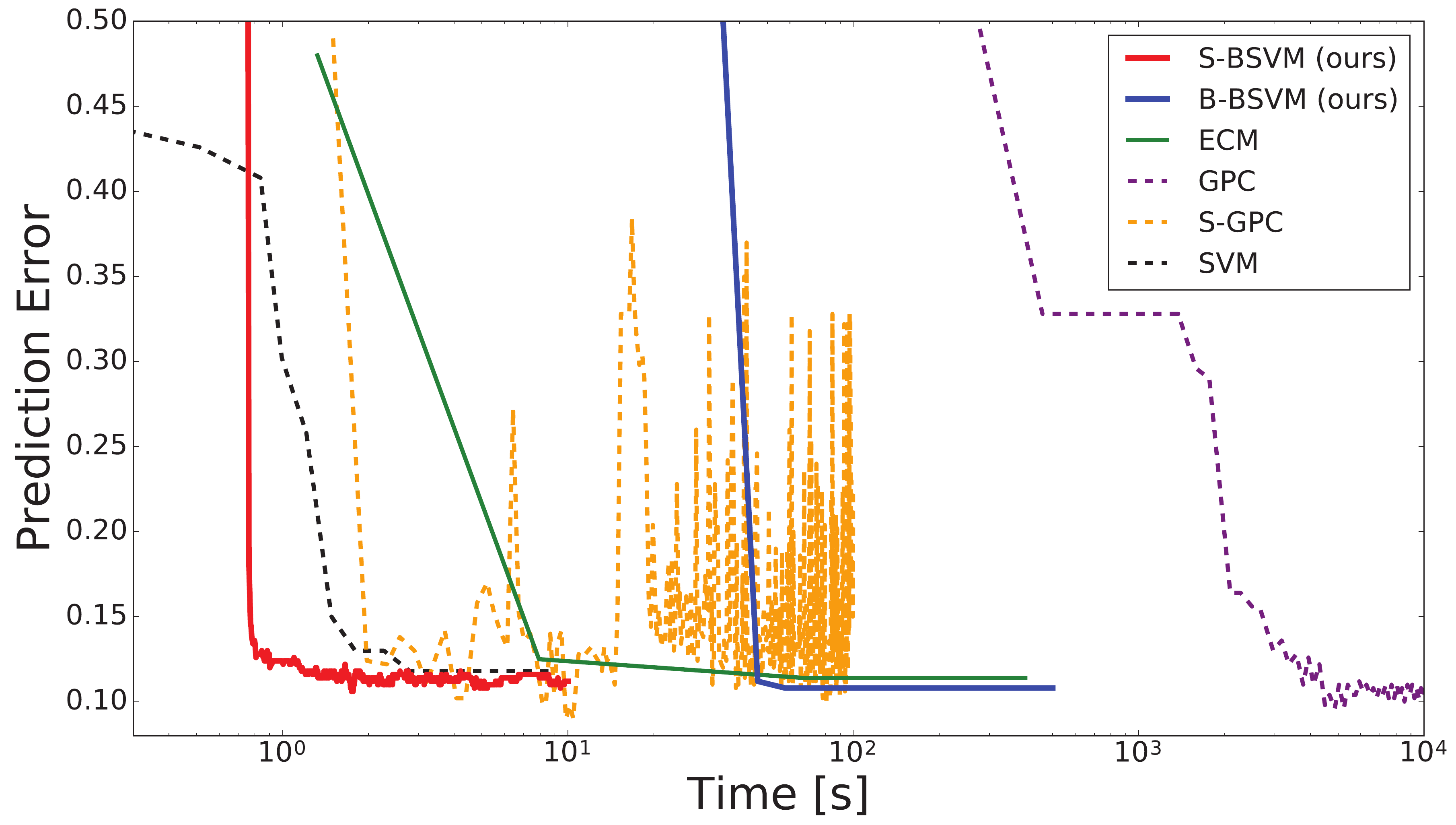}
\caption{Prediction error on held-out dataset vs. training time.}
\label{fig:timeexperiment}
\end{figure}

We examine the run time of our methods and the competitors. We include both the batch variational inference method (B-BSVM) described in section~\ref{sec:batchvi_non-linear} and our fast and scalable inference method (S-BSVM) described in section~\ref{sec:svi_non-linear} in the experiments. 
For each method we iteratively evaluate the prediction performance on a held-out dataset given a certain training time budget. The prediction error as function of the training time is shown in fig.~\ref{fig:timeexperiment}.
We experiment on the \emph{Waveform} dataset from the R\"{a}tsch benchmark dataset ($N=5000,\;d=21$). We use an RBF kernel with fixed length-scale parameter $\theta = 5.0$ and for the stochastic variational inference methods, S-BSVM and S-GPC, we use a batch size of 10 and 100 inducing points. 

Our scalable method (S-BSVM) is around 10 times faster than the direct competitor ECM while having slightly better prediction performance. The batch variational inference version (B-BSVM) is the slowest of the Bayesian SVM inference methods.
The related probabilistic model, Gaussian process classification, is around 5000 times slower than S-BSVM. Its stochastic inducing point version (S-GPC) has comparable run time to S-BSVM but is very unstable leading to bad prediction performance. S-GPC showed these instabilities for multiple settings of the hyperparameters.
The classic SVM (libSVM) has a similar run time as our method.
The speed and prediction performance of S-BSVM depend on the number of inducing points. See section~\ref{sec:exp_inducingpoints} for an empirical study. Note that the run time in table~\ref{tab:uncertainty} is determined after the methods have converged.
\subsection{Auto Tuning of Hyperparameters}
\begin{figure}
\centering
\includegraphics[scale=0.25]{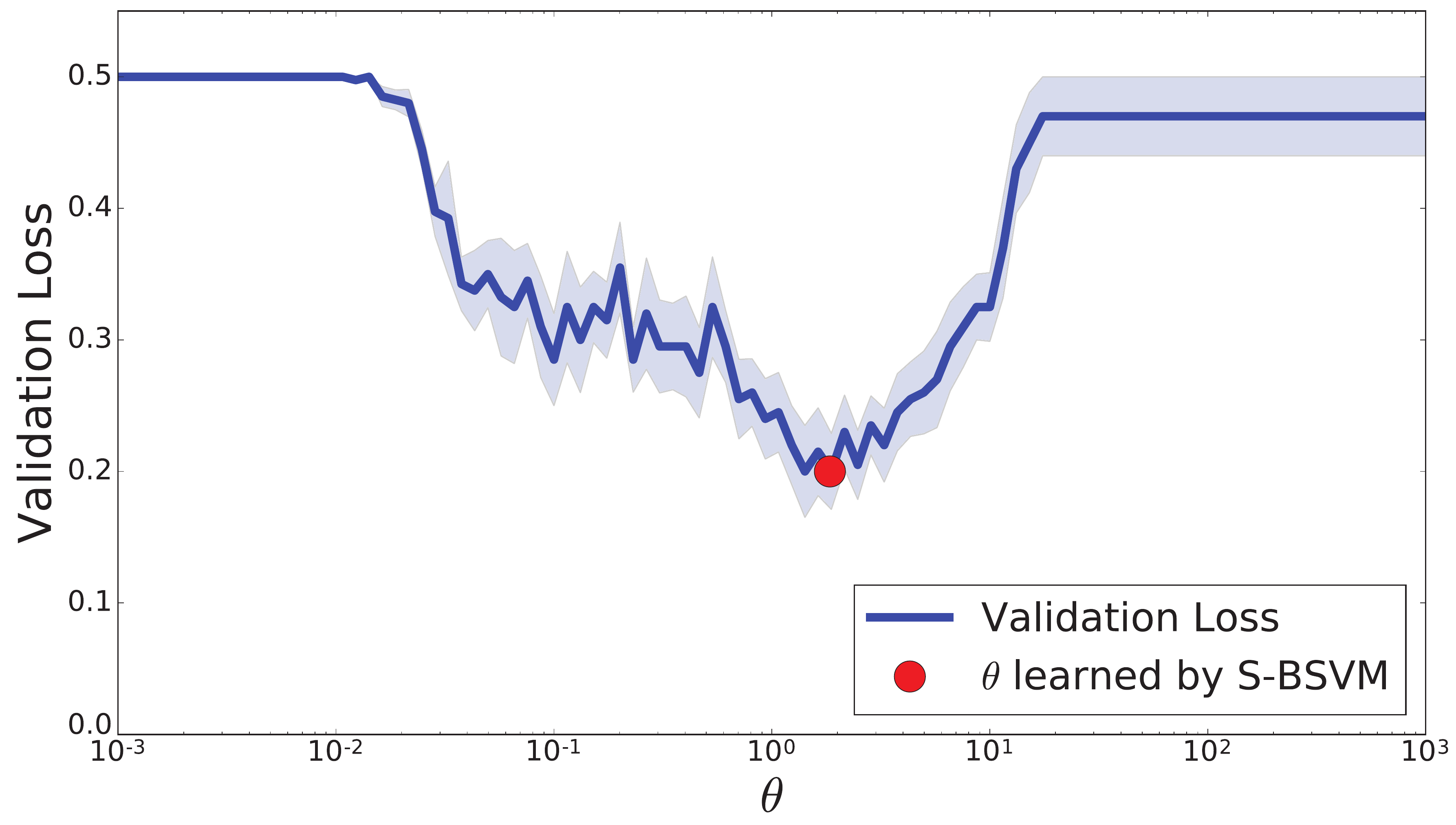}
\caption{Average validation loss as function of the RBF kernel length-scale parameter $\theta$, computed by grid search and 10-fold cross validation. The red circle represents the hyperparameter found by our proposed automatic tuning approach.}
\label{fig:autotuning}
\end{figure}

In section~\ref{sec:autotuning} we show that our inference method possesses the ability of automatic hyperparameter tunning. In this experiment we demonstrate that our method, indeed, finds the optimal length-scale hyperparameter of the RBF kernel.
We use the optimizing scheme \eqref{eq:hyperparam} and alternate between 10 variational parameter updates and one hyperparameter update.
We compute the true validation loss of the length-scale parameter $\theta$ by a grid search approach which consists of training our model (S-BSVM) for each $\theta$ and measuring the prediction performance using 10-fold cross validation. In fig.~\ref{fig:autotuning} we plot the validation loss and the length-scale parameter found by our method. We find the true optimum by only using 5 hyperparameter optimization steps. Training and hyperparameter optimization takes only 0.3 seconds for our method, whereas grid search takes 188 seconds (with a grid size of 1000 points).

\subsection{Inducing Points Selection}
\label{sec:exp_inducingpoints}
\begin{figure}
\begin{center}
\includegraphics[scale=0.25]{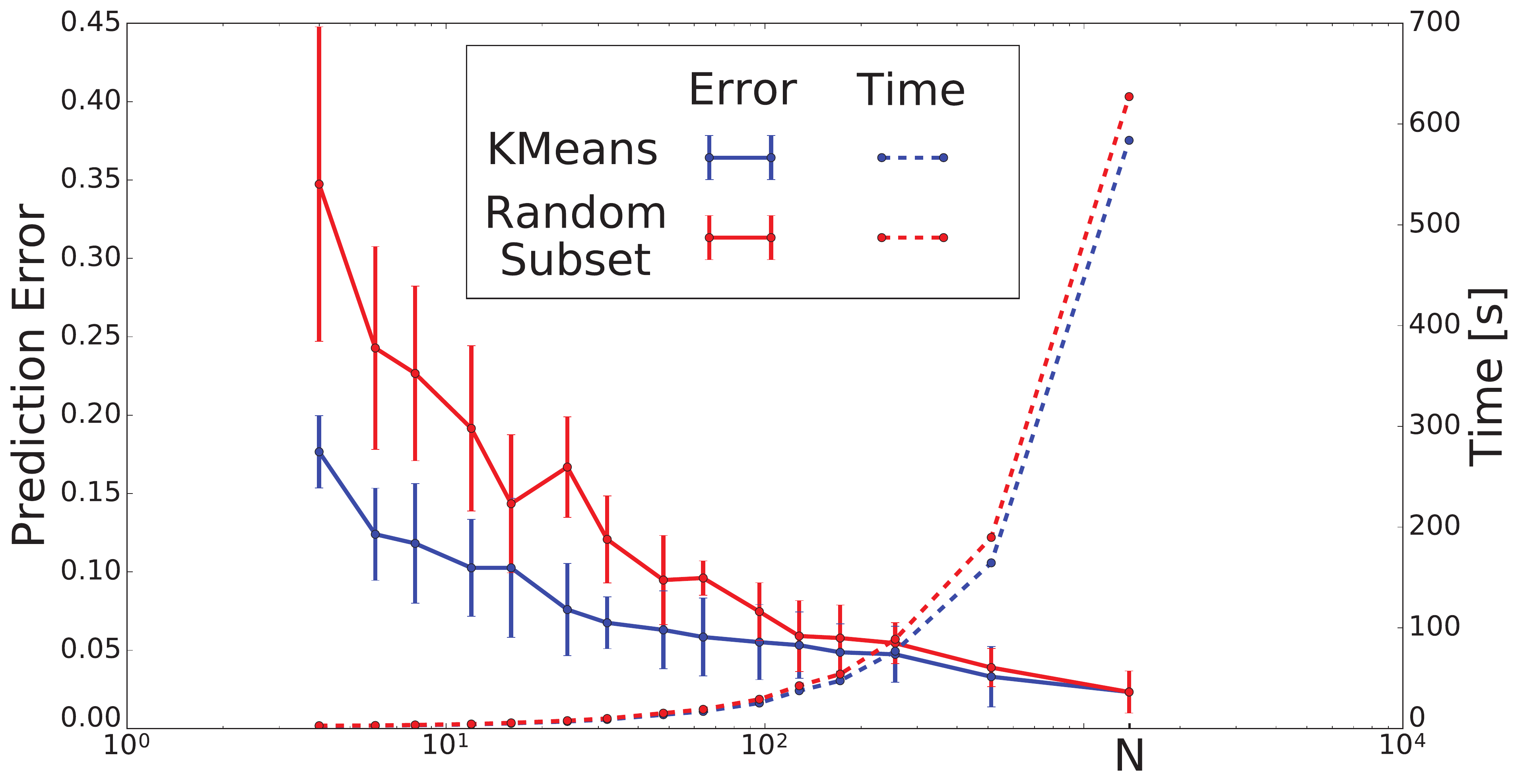}
\caption{Average prediction error and training time as functions of the number of inducing points selected by two different methods with one standard deviation (using 10-fold cross validation).}
\label{fig:inducingpoints}
\end{center}
\end{figure}

The sparse GP model used in our inference scheme builds on a set of inducing points where both the number and the locations of the inducing points are free parameters.
We investigate three different inducing point selection methods: random subset selection from the training set, the Gaussian Mixture Model (GMM), and the $k$-means clustering algorithm with an improved $k$-means++ seeding (kMeans) \cite{Bachem2016}. Furthermore we show how the number of inducing points affects the prediction accuracy and the run time.
We test the three inducing point selection methods on the USPS dataset \cite{UCI:2013} which we reduced to a binary problem using only the digits 3 and 5 (N=1350 and d=256).
For all methods we progressively increase the number of inducing points and compute the prediction error by 10-fold cross validation. We present our results in fig.~\ref{fig:inducingpoints}.
 
The GMM is unable to fit large numbers of samples and dimensions and fails to converge for almost all datasets tried, therefore, we do not include it in the plot.
Using the $k$-means selection algorithm leads for small numbers of inducing points to much better prediction performance than random subset selection.
Furthermore, we show that using only a small fraction of inducing points (around 1\% of the original dataset) leads to a nearly optimal prediction performance by simultaneously significantly decreasing the run time.
We observe similar results on all datasets we considered.

\section{Conclusion}
We presented a fast, scalable and reliable approximate inference method for the Bayesian nonlinear SVM. While previous methods were restricted to rather small datasets our method enables the application of the Bayesian nonlinear SVM to large real world datasets containing millions of samples. Our experiments showed that our method is orders of magnitudes faster than the state-of-the-art while still yielding comparable prediction accuracies. We showed how to automatically tune the hyperparameters and obtain prediction uncertainties which is important in many real world scenarios.

In future work we plan to further extend the Bayesian nonlinear SVM model to deal with missing data and account for correlations between data points building on ideas from \cite{probit2}. Furthermore, we want to develop Bayesian formulations of important variants of the SVM as for instance one-class SVMs \cite{conf/icdm/PerdisciGL06}.

\subsubsection*{Acknowledgments.}
We thank Stephan Mandt, Manfred Opper and Patrick Jähnichen for fruitful discussions. This work was partly funded by the German Research Foundation (DFG) award KL 2698/2-1.
\bibliographystyle{splncs}
\bibliography{refs}

\newpage
\appendix
\section{Appendix}

\subsection{Derivation of the Variational Objective}
In the following we give the details of the derivation of the variational objective \eqref{eq:ELBO_nonlin} for the inducing point model in section~\ref{sec:svi_non-linear}. The variational objective as defined in \eqref{eq:lowerbound_elbo} is
\begin{align*}
{\cal L} &= \mathbb{E}_q\left[\mathcal{L}_1\right] +  \mathbb{E}_q\left[\log p(u)\right] - \mathbb{E}_q\left[\log q(\lambda,u)\right]\label{eq:lowerbound_elbo}\\
			&= -\frac{1}{2}\sum_{i=1}^n\mathbb{E}_q\left[\log\lambda_i  + \frac{1}{\lambda_i}\left( \widetilde{K}_{ii}  + \left(1+\lambda_i - y_i K_{im}K_{mm}^{-1}u\right)^2\right)\right]\nonumber\\
			&\quad- \KL\left(q(u)||p(u)\right)  - \mathbb{E}_{q(\lambda)}\left[\log q(\lambda)\right].
\end{align*}
Using the abbreviation $\kappa_i = K_{im}K_{mm}^{-1}$ the first expectation term simplifies to
\begin{align*}
	&\mathbb{E}_q\left[\log\lambda_i  + \frac{1}{\lambda_i}\left( \widetilde{K}_{ii}  + \left(1+\lambda_i - y_i K_{im}K_{mm}^{-1}u\right)^2\right)\right]\\
    &= \mathbb{E}_q[\log\lambda_i] + \mathbb{E}_q\left[\lambda_i^{-1}\left(  \widetilde K_{ii} + 1 + \lambda_i^2 + \underbrace{y_i^2}_{=1}(\kappa_iu)^2 + 2\lambda_i - 2y_i\kappa_iu - 2\lambda_i y_i\kappa_iu \right)\right]\\
    &\uptoconst \mathbb{E}_{q(\lambda_i)}[\log\lambda_i] + \frac{1}{\sqrt{\alpha_i}} \left(  \widetilde{K}_{ii} + 1 + \lambda_i^2 + (\kappa_i\mu)^2 + \kappa_i \zeta \kappa_i^\top - 2y_i\kappa_i\mu \right) + \mathbb{E}_{q(\lambda_i)}[\lambda_i] - 2y_i \kappa_i \mu\\
    &= \mathbb{E}_{q(\lambda_i)}[\log\lambda_i] + \frac{1}{\sqrt{\alpha_i}} \left(  \widetilde{K}_{ii} + (1-y_i\kappa_i\mu)^2 + \lambda_i^2 +  \kappa_i \zeta \kappa_i^\top \right) + \mathbb{E}_{q(\lambda_i)}[\lambda_i] - 2y_i \kappa_i \mu.
\end{align*}
The entropy of $q(\lambda_i)$ is
\begin{align*}
\mathbb{E}_{q(\lambda_i)}\left[\log q(\lambda_i)\right] &= \mathbb{E}_{q(\lambda_i)}\left[ -\frac{1}{4}\log(\alpha_i)-\frac{1}{2}\log(\lambda_i) - \log(2)-\log(\mathrm{B}_\frac{1}{2}(\sqrt{\alpha_i}))-\frac{1}{2}\left(\lambda_i+\frac{\alpha_i}{\lambda_i}\right)\right]\\
		&\uptoconst  -\frac{1}{4}\log(\alpha_i)-\frac{1}{2}\mathbb{E}_{\alpha_i}\left[\log(\lambda_i)\right] -\log(\mathrm{B}_\frac{1}{2}(\sqrt{\alpha_i}))-\frac{1}{2}\mathbb{E}_{\alpha_i}\left[\lambda_i\right]-\frac{\alpha_i}{2}\mathbb{E}_{\alpha_i}\left[\frac{1}{\lambda_i}\right]\\
        &\uptoconst  -\frac{1}{4}\log(\alpha_i)-\frac{1}{2}\mathbb{E}_{\alpha_i}\left[\log(\lambda_i)\right] -\log(\mathrm{B}_\frac{1}{2}(\sqrt{\alpha_i}))-\frac{1}{2}\mathbb{E}_{\alpha_i}\left[\lambda_i\right] -\frac{\sqrt{\alpha_i}}{2},
\end{align*}
where $\mathrm{B}_{\frac{1}{2}}(.)$ is the modified Bessel function with parameter $\frac{1}{2}$ \cite{jorgensen2012statistical}.

By summing the terms the remaining expectations cancel out and we obtain
\begin{align*}
{\cal L} &\uptoconst  \sum_{i=1}^n \Big\{ -\frac{1}{2} \mathbb{E}_{q(\lambda_i)}[\log\lambda_i] - \frac{1}{2\sqrt{\alpha_i}} \left(  \widetilde{K}_{ii} + (1-y_i\kappa_i\mu)^2 + \lambda_i^2 +  \kappa_i \zeta \kappa_i^\top \right)   -\frac{1}{2} \mathbb{E}_{q(\lambda_i)}[\lambda_i] + y_i \kappa_i \mu\\
	&\quad +   \frac{1}{4}\log(\alpha_i)+\frac{1}{2}\mathbb{E}_{q(\lambda_i)}\left[\log(\lambda_i)\right] +\log(\mathrm{B}_\frac{1}{2}(\sqrt{\alpha_i})) +\frac{1}{2}\mathbb{E}_{q(\lambda_i)}\left[\lambda_i\right]  +\frac{\sqrt{\alpha_i}}{2} \Big\} - \KL\left(q(u)||p(u)\right)\\
	&= \sum_{i=1}^n \Big\{ - \frac{1}{2\sqrt{\alpha_i}}\left(  \widetilde{K}_{ii} + (1-y_i\kappa_i\mu)^2 + \lambda_i^2 +  \kappa_i \zeta \kappa_i^\top - \alpha_i \right)  + y_i \kappa_i \mu +   \frac{1}{4}\log(\alpha_i) +\log(\mathrm{B}_\frac{1}{2}(\sqrt{\alpha_i}))  \Big\}\\
    &\quad - \KL\left(q(u)||p(u)\right)\\
    &\uptoconst \sum_{i=1}^n \Big\{ - \frac{1}{2\sqrt{\alpha_i}} \left(  \widetilde{K}_{ii} + (1-y_i\kappa_i\mu)^2 + \lambda_i^2 +  \kappa_i \zeta \kappa_i^\top - \alpha_i\right)  + y_i \kappa_i \mu +   \frac{1}{4}\log(\alpha_i) +\log(\mathrm{B}_\frac{1}{2}(\sqrt{\alpha_i}))  \Big\}\\
    &\quad + \frac{1}{2} \log |\zeta| - \frac{1}{2}\text{tr}(K_{mm}^{-1}\zeta)-\frac{1}{2}\mu^\top K_{mm}^{-1}\mu\\
    &= \frac{1}{2} \log |\zeta| - \frac{1}{2}\text{tr}(K_{mm}^{-1}\zeta)-\frac{1}{2}\mu^\top K_{mm}^{-1}\mu + y^\top\kappa\mu\\
		&\quad +\sum_{i=1}^n\left\{\log(\mathrm{B}_{\frac{1}{4}}(\sqrt{\alpha_i})) + \frac{1}{2}\log(\alpha_i)-\frac{1}{2}\alpha_i^{-\frac{1}{2}}\left( 1 - \alpha_i - 2y_i\kappa_{i.}\mu + \left(\kappa(\mu\mu^\top + \zeta) \kappa^\top + \widetilde{K}\right)_{ii}  \right)\right\}.
\end{align*}

\subsection{Euclidean and Natural Gradients of the Variational Objective}
First, we compute the standard euclidean gradients of ${\cal L}$. The derivative w.r.t. the mean and covariance matrix are
\begin{align*}
		\der{{\cal L}}{\zeta} &= \frac{1}{2}\left(\deralt{\log\vert\zeta\vert}{\zeta} - \deralt{\text{tr}(K_{mm}^{-1}\zeta)}{\zeta}\right) + \sum_{i=1}^N -\frac{1}{2\sqrt{\alpha_i}}\deralt{y_i\kappa_i\zeta\kappa_i^\top y_i}{\zeta}\\
		&= \frac{1}{2}\left(\zeta^{-1}\right)^T - \frac{1}{2}\left(K_{mm}^{-1}\right)^T - \frac{1}{2}Y^2\kappa^\top A^{-\frac{1}{2}}\kappa\\
		&= \frac{1}{2}\left(\zeta^{-1} - K_{mm}^{-1} - \kappa^\top A^{-\frac{1}{2}} \kappa \right)\\
        &=: {\cal L}'_\zeta,
\end{align*}
with $A = \text{diag}(\alpha)$ and
		\begin{align*}
		\der{{\cal L}}{\mu} &= -\frac{1}{2}\deralt{\mu^TK_{mm}^{-1}\mu}{\mu} + \sum_{i=1}^N \deralt{y_i\kappa_i \mu}{\mu} + \frac{1}{2\sqrt{\alpha_i}}\deralt{(1-y_i\kappa_i\mu)^2}{\mu}\\
		&= -K_{mm}^{-1}\mu + \sum_{i=1}^N y_i\kappa_i + \frac{1}{\sqrt{\alpha_i}}\left(y_i\kappa_i^\top - y_i^2\kappa_i^\top\kappa_i\mu\right) \\
		&= -K_{mm}^{-1}\mu + \kappa^\top y + \kappa^\top Y\alpha^{-\frac{1}{2}} + \kappa^\top A^{-\frac{1}{2}}\kappa\mu\\
		&= -\left(K_{mm}^{-1} + \kappa^\top A^{-\frac{1}{2}}\kappa\right)\mu + \kappa^\top Y(\alpha^{-\frac{1}{2}}+1)\\
        &=: {\cal L}_\mu'.
\end{align*}
The derivative w.r.t. parameter $alpha_i$ of the generalized inverse Gaussian distribution is
\begin{align*}
		\der{{\cal L}}{\alpha_i} &= \frac{1}{4}\deralt{\log(\alpha_i)}{\alpha_i} + \deralt{\log(K_{\frac{1}{2}}(\sqrt{\alpha_i}))}{\alpha_i} + \frac{1}{2}\deralt{\sqrt{\alpha_i}}{\alpha_i} - \frac{(1-y_i\kappa_i\mu)^2 + y_i(\kappa_i\zeta\kappa_i^\top + \widetilde{K}_{ii}) y_i}{2}\deralt{\frac{1}{\sqrt{\alpha_i}}}{\alpha_i}\\
		&= \frac{1}{4\alpha_i}-(\frac{1}{4\alpha_i}+\frac{1}{2\sqrt{\alpha_i}})+\frac{1}{4\sqrt{\alpha_i}}+\frac{(1-y_i\kappa_i\mu)^2 + y_i(\kappa_i\zeta\kappa_i^\top + \widetilde{K}_{ii})y_i}{4\sqrt{\alpha_i}^3}\\
		&= \frac{(1-y_i\kappa_i\mu)^2 + y_i(\kappa_i\zeta\kappa_i^\top + \widetilde{K}_{ii})y_i}{4\sqrt{\alpha_i}^3} - \frac{1}{4\sqrt{\alpha_i}}.
	\end{align*}        
The natural gradient can be computed by pre-multiplying the euclidean gradient with the inverse Fisher information matrix \cite{infogeom}. Applied to a Gaussian distribution this leads to the following expressions for the natural gradient w.r.t. the natural parameters \cite{infogeom},
\begin{align*}
\widetilde{\nabla}_{(\eta_1, \eta_2)} {\cal L}(\eta) = \left({\cal L}_\mu'(\eta) - 2{\cal L}_\zeta'(\eta)\mu,\; {\cal L}_\zeta'(\eta)\right).
\end{align*}	
Using the identities $\eta_1 = \zeta^{-1}\mu$ and $\eta_2 = -\frac{1}{2}\zeta^{-1}$ we obtain
\begin{align*}
{\cal L}_\mu'(\eta) &= \frac{1}{2}\left(K_{mm}^{-1} + \kappa^\top A^{-\frac{1}{2}}\kappa\right)\eta_2^{-1}\eta_1 + \kappa^\top Y(\alpha^{-\frac{1}{2}}+1)\\
{\cal L}_\zeta'(\eta) &= \frac{1}{2}\left(-2\eta_2 - K_{mm}^{-1} - \kappa^\top A^{-\frac{1}{2}} \kappa \right) = -\frac{1}{2}(K_{mm}^{-1} + \kappa^\top A^{-\frac{1}{2}} \kappa) - \eta_2
\end{align*}	
Finally, this leads to the natural gradients with respect to the natural parameters,     
\begin{align*}
	\widetilde{\nabla}_{\eta_1} {\cal L} &= {\cal L}_\mu' - 2{\cal L}_\zeta'\mu\\
    &= \frac{1}{2}\left(K_{mm}^{-1} + \kappa^\top A^{-\frac{1}{2}}\kappa\right)\eta_2^{-1}\eta_1 + \kappa^\top Y(\alpha^{-\frac{1}{2}}+1) + \frac{1}{2}\left(-2\eta_2 - K_{mm}^{-1} - \kappa^\top A^{-\frac{1}{2}} \kappa \right)\eta_2^{-1}\eta_1\\
&= \kappa^\top Y(\alpha^{-\frac{1}{2}}+1) - \eta_1,
\intertext{and}
	\widetilde{\nabla}_{\eta_2} {\cal L} &= {\cal L}_\zeta'= -\frac{1}{2}(K_{mm}^{-1} + \kappa^\top A^{-\frac{1}{2}} \kappa) - \eta_2.
\end{align*}
        
\subsection{Optimization of the Kernel Hyperparameters}
We consider a general multiple kernel approach. Let  $k(x,x') = \sum_j\gamma_j k_j(x,x',\theta_j)$ be the kernel function where  $\theta_j$ denote the hyperparameters of the kernel function $k_j$ (e.g. the length scale parameter of an RBF kernel) and $\gamma_j$ the corresponding kernel weight.
Let $\omega = \{\theta_j, \gamma_j\}_{j=1,...,J}$ be the collection of all hyperparameters. The derivative of the variational objective ${\cal L}$ w.r.t. to the hyperparameters is

\begin{align*}
\der{L}{\omega} &= -\frac{1}{2}\deralt{}{\omega}\Big(\log\vert K_{mm}\vert + \text{tr}(K_{mm}^{-1}\zeta)+\mu^T K_{mm}^{-1}\mu - 2(1+\alpha^{-\frac{1}{2}^\top})Y\kappa\mu\\
&\quad +\alpha^{-\frac{1}{2}^\top}\diag\left(\kappa(\mu\mu^\top + \zeta) \kappa^\top + \widetilde{K}\right)\Big)
\end{align*}
Using the abbreviations  $J_{**} ^\omega = \frac{\mathrm{d}K_{**}}{\mathrm{d}\omega}$ and $\iota^\omega = \frac{\mathrm{d}\kappa}{\mathrm{d}\omega} = \left(J_{nm}^\omega - \kappa J_{mm}^\omega \right)K_{mm}^{-1}$ we obtain
\begin{align*}
\der{L}{\omega} &= -\frac{1}{2}\left(\Tr\left(K_{mm}^{-1}J_{mm}^\omega\left(\mathbb{I}-K_{mm}^{-1}\zeta\right)\right) - \left(\mu^\top K_{mm}^{-1}J_{mm}^\omega K_{mm}^{-1} + 2(1+\alpha^{-\frac{1}{2}^\top})Y\iota^\omega\right)\mu \right.\\
&+ \left. \alpha^{-\frac{1}{2}^\top}\text{diag}\left[\kappa\left(\left(\mu\mu^\top + \zeta\right){\iota^\omega}^\top - J^\omega_{mn}\right) + \iota^\omega\left(\left(\mu\mu^\top + \zeta\right)\kappa^\top - K_{mn}\right) + J_{nn}^\omega \right]\right). 
\end{align*}
To compute the gradient w.r.t. to specific hyperparameters we only have to plug in the derivatives of the kernel function $\frac{\mathrm{d}K_{**}}{\mathrm{d}\omega}$ into the above formula.

\end{document}